\documentclass{article}

\PassOptionsToPackage{numbers,sort&compress}{natbib}
\usepackage[preprint]{2026_neurips}

\usepackage[utf8]{inputenc}
\usepackage[T1]{fontenc}
\usepackage{courier}
\usepackage{hyperref}
\usepackage{url}
\usepackage{booktabs}
\usepackage{amsmath}
\usepackage{amsfonts}
\usepackage{graphicx}
\usepackage{microtype}
\usepackage{xcolor}
\usepackage{tabularx}

\hypersetup{
  pdfauthor={Qinzhen Ma and Jialin Wu},
  pdftitle={Guardrailed Meta-Agent Loops: Stress-Testing Policy Pinning, Budget Bounds, and Crash Recovery},
  pdfsubject={Preprint},
  pdfcreator={LaTeX},
  pdfproducer={pdfTeX},
  colorlinks=true,
  linkcolor=blue,
  citecolor=blue,
  urlcolor=blue
}
\title{Guardrailed Meta-Agent Loops:\\
Stress-Testing Policy Pinning, Budget Bounds, and Crash Recovery}
\author{Qinzhen Ma\\Rice University\\\texttt{qm18@rice.edu}\And Jialin Wu\\University of California, San Diego\\\texttt{jlwu@ucsd.edu}}

\newcommand{\sys}{\textsc{GuardrailLoop}}

\newcommand{\gphs}{GPU-hours}

\begin{document}
\maketitle

\begin{abstract}
Self-improving agent workflows create an audit problem when the same controller can change both its behavior and the conditions under which that behavior is judged. We present \sys, a simulation-based testbed that makes three operational contracts jointly testable: preservation of human-defined policy, compute accounting at every recorded execution prefix, and recovery of a specified scientific state after crashes. A hash-pinned policy fixes goals, scope, evaluation identity, budget, and release conditions; machine-directed evolution is restricted to a code-owned feature catalog and bounded knobs. The contribution is an executable boundary and an evaluation protocol that separates useful adaptation, state recovery, and repeated execution. In a paired 50-seed $2\!\times\!2$ study, round-stage growth changes target attainment by $+1.00$ and restricted mean compute to target by $-56.97$ simulated \gphs{} (95\% paired-bootstrap interval $[-58.91,-54.70]$); idle growth has zero measured utility effect. Across 240 enumerated crash injections, all runs recover the defined outcome, but only 210 preserve the normalized trace: 30 pre-commit crashes repeat a planner call. Resource-drift, kill-switch, integrity, and output-guard matrices satisfy their specified checks. These findings show why successful outcome recovery is insufficient evidence of exactly-once execution. They establish conformance within one calibrated deterministic testbed, rather than general safety or real-world self-improvement.
\end{abstract}

\section{Introduction}
\label{sec:introduction}

An agent that manages other agents determines which work is attempted, how data and compute are allocated, which checkpoint is evaluated, and whether another iteration begins. If the controller also revises the goal, stopping rule, or evaluation identity, an apparent improvement can reflect a changed criterion rather than better behavior. Persistent workflows add a second difficulty: a crash or resource change can alter what is executed, charged, and replayed even when the final checkpoint appears correct. The central question is therefore operational: \emph{what may improve, what must remain fixed, and what evidence is needed after execution is interrupted?}

We study this question through three distinct contracts. \emph{Authority preservation} limits machine-directed writes while retaining human control of the run policy. \emph{Prefix accounting} checks the resource bound throughout the recorded history, including unfinished work. \emph{Recovery equivalence} specifies which parts of state must agree after restart and separately records whether requests were repeated. These contracts address different failure surfaces: a valid final budget total does not by itself establish a bound at every earlier prefix, and a matching final outcome does not establish identical execution.

\sys{} instantiates this specification in a fixed five-role orchestrator with a hash-pinned human \emph{policy plane} and a bounded machine \emph{evolution plane}. The planner can select features from a closed, code-owned catalog but cannot supply the numeric effect of a feature or write policy fields. Goals, task scope, evaluation identity, budget, and release conditions remain fixed for the run. The six-stage loop combines these interfaces with verified replay, per-step snapshots, atomic evolution commits, and accounting for in-flight work. New-agent suggestions remain inert records; the prototype does not recursively deploy agents.

Our contribution is the concrete coupling of these contracts and their evidence, rather than a new hash-chain primitive or a general safety guarantee. We test whether permitted evolution can affect utility while the authority, accounting, and recovery checks remain separately observable. This separation matters in the results: round growth is useful in the calibrated simulator, idle growth is not, and all 240 crash cases preserve the specified scientific outcome although 30 repeat a planner call. The latter finding makes the gap between state recovery and repeated execution a first-class result.

\paragraph{Contributions.}
\begin{enumerate}
\item \textbf{An explicit boundary for bounded adaptation.} We specify the state that human policy pins and the code-mediated changes available to the machine, including the separation between selecting a feature and defining its effect.
\item \textbf{A joint accounting and recovery specification.} We connect durable state transitions to prefix compute accounting and define scientific-state equivalence separately from normalized-trace identity, exposing the scope of a successful restart.
\item \textbf{A paired utility study and enumerated conformance tests.} Growth ablations, a budget sweep, and crash, drift, integrity, kill, and output-guard matrices distinguish useful adaptation from operational correctness within the stated simulator and fault model.
\end{enumerate}

\section{Related work and scope of the contribution}
\label{sec:related}

\paragraph{Self-improvement and the object of adaptation.}
G\"odel machines study self-referential improvement under a formal objective~\citep{schmidhuber2007godel}. STOP recursively improves a code optimizer, ADAS searches agent programs, and G\"odel Agent and the Darwin G\"odel Machine explore modification of agent implementations~\citep{zelikman2024stop,hu2025adas,yin2025godelagent,zhang2026dgm}. These works motivate asking what an improving system may change. Our adaptation space is deliberately narrower: catalog identifiers are resolved to code-owned configuration changes inside a fixed orchestrator. This restriction makes authority and recovery behavior tractable to inspect, but also limits the scope of any learning claim. The experiments do not compare performance with open-ended code-search systems.

\paragraph{Runtime constraints and privileges.}
AgentSpec expresses behavioral restrictions through declarative triggers, predicates, and enforcement rules~\citep{wang2026agentspec}. Progent checks tool privileges; its policy-update mechanism distinguishes permission narrowing from expansion and routes expansion to a configurable approver~\citep{li2025progent}. Thus deterministic enforcement and controlled policy updates are established ideas. The particular boundary studied here pins experimental governance---goal, exit condition, evaluation identity, budget, scope, and release---and permits catalog-bounded changes to learning configuration. We evaluate this boundary together with accounting and recovery, without claiming that it provides stronger security than prior enforcement frameworks.

\paragraph{Resource-constrained planning.}
INTENT studies budget-constrained tool use with priced, stochastic executions and planning under cost uncertainty~\citep{xu2026intent}. Our resource variable is simulated training/evaluation compute, and the question is whether recorded settled and unfinished charges remain within a pinned cap when execution is interrupted or repriced. This is an accounting contract, not a method for learning the value of a tool call or predicting its monetary cost. Both perspectives concern budget feasibility but at different execution interfaces.

\paragraph{Durable execution and recovery observables.}
Durable Functions gives formal semantics and equivalence results for record-and-replay execution~\citep{burckhardt2021durable}. AgentRewind restores aligned agent context and controlled-environment state to support revised continuations after errors~\citep{zhang2026agentrewind}. We examine continuation after enumerated infrastructure crash points and report scientific-state equality separately from normalized-trace equality. This follows the broader motivation to decompose reliability rather than collapse it into task success~\citep{kapoor2026science}. The contribution is a concrete run specification and its conformance evidence for bounded adaptation, including the observed separation between outcome recovery and repeated planner calls. It is not a new general replay theorem or an exactly-once guarantee for external effects.

\section{Guardrailed meta-agent testbed}
\label{sec:testbed}

\subsection{Execution and enforcement mechanisms}

\paragraph{Control and evolution planes.}
At creation, the system freezes a policy containing the goal; target, budget, and maximum rounds; a task whitelist; evaluation-set identifier and hash; release thresholds; seed; model kind; and owner. The policy hash and evaluation hash are recorded in the run history and checked before every stage. Operator settings such as GPU count and step duration may change on resume but cannot redefine success. Recursive, case-insensitive key filters and stage-specific schemas reject policy-shaped model outputs. These checks constrain implemented interfaces, not a privileged process with filesystem access.

\begin{table}[h]
\caption{The authority boundary is asymmetric: generated identifiers are interpreted by trusted code, while the policy has no model-output write path.}
\label{tab:planes}
\centering
\small
\begin{tabularx}{\linewidth}{@{}>{\bfseries}l X X@{}}
\toprule
 & Human policy plane (hash-pinned) & Machine evolution plane (bounded) \\
\midrule
State & goal, exit/budget, scope, eval hash, release, seed/model/owner & task/data choices; feature IDs $\rightarrow$ seven knobs + self-check \\
Authority & verified before each stage; code-only exit judge & $\leq$2 features/round; $\leq$1/idle; proposals inert \\
\bottomrule
\end{tabularx}
\end{table}

The evolvable state is deliberately smaller. A model feature-pick output contains only a catalog identifier, rationale, and evidence. Trusted code constructs the Feature, supplies its deltas, and clamps seven optimization/data knobs plus a self-check flag. Task cards are filtered against the whitelist at schema, guard, data, and simulator layers. Eval receives neither knobs nor an LLM client and runs the pinned evaluation set. A pure-code judge is the sole constructor of exit decisions. The orchestrator executes \texttt{DECOMPOSE $\rightarrow$ DATA $\rightarrow$ TRAIN $\rightarrow$ EVAL $\rightarrow$ JUDGE $\rightarrow$ EVOLVE}, with optional idle reflection. Even with both growth paths disabled, Task and Data still react to prior failures, so our control is \emph{no Feature-growth}, not a static learner.

\paragraph{Recovery and audit semantics.}
Each durable transition is appended to a hash-linked JSONL ledger; replay first verifies the full chain and then folds events into state. A torn final record may be truncated before append, whereas an edited or deleted complete prefix fails closed. This is tamper evidence, not tamper proof: the chain has no signature or external anchor, so a writer able to rewrite and re-hash the full history is outside the modeled threat.

Training emits a snapshot and cumulative charge at every step. Under unchanged execution settings, recorded RNG coordinates support deterministic continuation. Round evolution is installed by one durable \texttt{stage\_done} event; idle problems, features, proposals, and knobs are installed by one \texttt{idle\_finished} event. A pre-commit crash can repeat the planner call, hence calls are \emph{at least once}; only the recorded state transition is atomic. Outcome equivalence therefore depends on the deterministic offline planner evaluated here.

At every ledger prefix, the accounting check is $H_{\mathrm{settled}}+H_{\mathrm{inflight}}\leq H_{\mathrm{policy}}+\epsilon$, with numerical tolerance $\epsilon=10^{-9}$. On resume with different GPUs or step length, the orchestrator first reserves evaluation cost and reprices affordable unfinished work from the exact recorded charge. This supports a ledger-accounted simulated-compute bound; it neither meters a cloud bill nor prevents unrecorded physical work after power loss.

\subsection{Operational contracts and observable evidence}
\label{sec:contracts}

At a recorded boundary $k$, write the execution state as $S_k=(P_k,M_k,W_k,L_k;u_k)$: $P_k$ is the human policy, $M_k$ the mutable evolution state, $W_k$ the training/evaluation and orchestration state, and $L_k$ the durable ledger prefix. Operator settings $u_k$, such as GPU count and step duration, may change on resume but do not grant the machine a write path to $P_k$. This notation summarizes the implemented interfaces and tested predicates; it is not a machine-checked proof of the implementation.

\paragraph{Authority contract.}
For accepted machine-directed transitions, the intended relation is $P_{k+1}=P_k=P_0$. Feature selection changes only the allowed evolution state. For a numeric knob vector $\theta$ and selected catalog feature $f$, the update has the form
\begin{equation}
 f\in\mathcal{F},\qquad
 \theta' = \operatorname{clip}_{\mathcal{K}}
 \bigl(\theta+\Delta_{\mathrm{code}}(f)\bigr),
 \label{eq:authority}
\end{equation}
where both the catalog $\mathcal{F}$ and permitted ranges $\mathcal{K}$ are code-owned. The self-check flag and task/data choices have their own bounded interfaces. A model may choose an admissible operation; it does not choose that operation's numeric authority. Hash checks detect disagreement with recorded policy and evaluation identities within the stated localized-corruption model. They do not protect against a privileged writer who can replace and re-hash the entire history.

\paragraph{Accounting contract.}
For every recorded ledger prefix, the monitored predicate is
\begin{equation}
 H_{\mathrm{settled}}(k)+H_{\mathrm{inflight}}(k)
 \leq B(P_0)+\epsilon,\qquad \epsilon=10^{-9},
 \label{eq:budget}
\end{equation}
where $B(P_0)$ is the policy budget in simulated \gphs. A resume under a changed resource configuration must price unfinished work from the recorded cumulative charge and preserve the evaluation reservation. The predicate concerns ledger-accounted work, so it cannot establish a bound on physical work that was never durably recorded.

\paragraph{Recovery contract.}
Let $\Pi(S)$ retain the scientific outcome specified in Appendix~\ref{app:faults}: decision, accounted compute/time, evaluation curve, checkpoint identities, and Feature/proposal/problem state. It excludes the full ledger and planner-call trace; ledger validity is checked separately. Let $\mathcal{T}$ denote the execution trace after removing wall-clock, latency, and expected recovery bookkeeping. For seed $s$ and crash site $c$, define two separate checks:
\begin{align}
 R_{s,c} &= \mathbf{1}\!\left\{
 \Pi(S_s^{\mathrm{base}})=\Pi(S_{s,c}^{\mathrm{resume}})\right\},
 \label{eq:recovery}\\
 T_{s,c} &= \mathbf{1}\!\left\{
 \mathcal{T}_s^{\mathrm{base}}=\mathcal{T}_{s,c}^{\mathrm{resume}}\right\}.
 \label{eq:trace}
\end{align}
The reported recovery suite evaluates these checks with an offline deterministic planner and unchanged execution settings. Resource drift is a separate accounting test. Across its 240 crash cells, $R_{s,c}=1$ in every case while $T_{s,c}=1$ in only 210. A repeated deterministic call can leave $\Pi(S)$ unchanged while changing $\mathcal{T}$. Thus the evaluated recovery contract intentionally exposes an execution-level difference that final-state equality alone would conceal. Appendix~\ref{app:analysis} gives conditional arguments for the admission bound and the limit of local commit atomicity.

\begin{table}[t]
\caption{Contracts, reported evidence, and the boundary of each claim. Counts describe the enumerated experiments, not probabilities of deployment failure.}
\label{tab:contracts}
\centering
\small
\begin{tabularx}{\linewidth}{@{}l X X@{}}
\toprule
Contract & Reported witness & Limit of the witness \\
\midrule
Authority & 176 output-guard cases; 70 complete-artifact corruptions rejected & Coverage of specified inputs and localized corruption; no full-history rewrite resistance \\
Accounting & Prefix checks in 200 budget runs and 40 resource-drift runs & Simulated ledger charge; no physical or monetary metering \\
Recovery & 240/240 scientific outcomes; 210/240 normalized traces & Deterministic planner and enumerated crash sites; repeated calls remain observable \\
\bottomrule
\end{tabularx}
\end{table}

\section{Evaluation}
\label{sec:evaluation}

\paragraph{Questions and evidence.}
The experiments answer three separate questions: does catalog-bounded growth affect utility (paired growth ablation); does accounting remain within policy as the budget or execution configuration changes (budget sweep and drift suite); and which recovery and authority checks hold at the enumerated boundaries (fault and guard matrices)? Table~\ref{tab:contracts} maps the contracts to their witnesses. The utility ablation varies growth availability, not the presence of guards or recovery mechanisms, so it does not estimate the performance cost or necessity of each enforcement component.

\paragraph{Protocol.}
The single-skill simulator contains one simulated VLA and five shirt-folding tasks. Each pinned evaluation uses 1,000 Bernoulli episodes (200 per task). The planner is an offline deterministic rule table; seeds affect initialization, training noise, and evaluation draws. Unless varied, the target is 0.95, budget 160 simulated \gphs, round time box four hours on four GPUs, and step length 0.25 hours. The experimental unit is a complete seed-run, not a task, round, or episode.

For utility, 50 seeds are paired across a $2\!\times\!2$ design: round-stage growth $E\in\{0,1\}$ and idle reflection/growth $I\in\{0,1\}$. Primary outcomes are observed target attainment and restricted mean \gphs{} to target, with non-hits censored at 160. We also record the selected (best-observed) checkpoint's observed score and that checkpoint's noise-free latent simulator score. Continuous means and factorial contrasts use 10,000 seed-block bootstrap replicates; single proportions use Wilson intervals. A second paired sweep varies policy budget over $\{20,40,80,160\}$.

\begin{figure}[t]
  \centering
  \includegraphics[width=\linewidth]{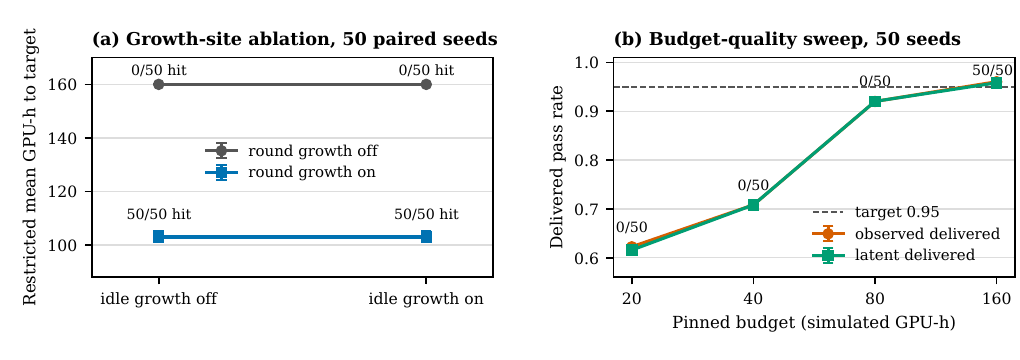}
  \caption{\textbf{Utility in the calibrated simulator.} (a) Round-stage growth reduces restricted mean compute to target from 160 to 103.03 \gphs{} and changes observed hits from 0/50 to 50/50; idle growth has no measured utility effect here. Error bars are 95\% seed-bootstrap intervals. (b) Selected-checkpoint observed and latent scores rise with the pinned budget; labels give observed hits. The 0.95 threshold is crossed only at budget 160.}
  \label{fig:results}
\end{figure}

\paragraph{Growth and budget results.}
Full and round-only runs are identical on the reported utility outcomes: both hit the observed target in 50/50 seeds (Wilson 95\% interval $[0.929,1]$), use 103.03 mean \gphs{} ($[101.09,105.30]$), and deliver latent pass rate 0.9584 ($[0.9564,0.9605]$). Idle-only and no-growth each hit 0/50, consume the 160-hour censoring budget, and deliver latent rate 0.8324 ($[0.8290,0.8362]$). Thus the paired round-growth effects are $+1.00$ on target attainment, $-56.97$ \gphs{} on restricted mean compute ($[-58.91,-54.70]$), and $+0.1260$ on latent score ($[0.1220,0.1295]$). Idle and interaction contrasts are exactly zero in these runs. This negative result is specific to the present catalog and dynamics, not evidence that idle reflection is generally useless.

At budgets 20, 40, 80, and 160, target hits are respectively 0, 0, 0, and 50 of 50; mean latent delivered rates are 0.6156, 0.7082, 0.9203, and 0.9584. Across the 200 budget runs, 13 deliver a best-observed checkpoint other than the last, exposing selection effects. The maximum recorded prefix excess is $1.42\!\times\!10^{-14}$ \gphs{} (floating-point error; tolerance $10^{-9}$), and maximum settlement reconciliation error is $2.84\!\times\!10^{-14}$. These are internal ledger checks, not independent metering.

\paragraph{Fault matrices.}
For each of 10 seeds we compare a two-round uninterrupted baseline with 24 crash sites: before, pre-commit, and after all six stages; training ticks 1/5/12; and idle start/pre-commit/completion. We separately test four GPU/step-size drift paths, kill requests at ticks 1/5/12, seven complete-artifact corruptions plus a torn tail, and a deterministic 176-case output-guard matrix (19 protected keys $\times$ three casings $\times$ three nestings, plus five catalog/scope/authority cases). Table~\ref{tab:faults} reports exhaustive outcomes over these enumerated cells, not IID reliability estimates.

\begin{table}[h]
\caption{All specified checks pass, but normalized traces expose repeated pre-commit planner calls. ``Budget'' is simulated ledger accounting.}
\label{tab:faults}
\centering
\small
\begin{tabularx}{\linewidth}{@{}l r X@{}}
\toprule
Stress & Result & Exact scope of passing check \\
\midrule
Crash injection & 240/240 & decision, checkpoints, curve, projected scientific state, GPU/sim time, chain \\
Normalized trace & 210/240 & 30 mismatches: one replayed call at each of 3 pre-commit sites/seed \\
Resource drift & 40/40 & every prefix within 20 \gphs; monotone tick gauge; eval completes \\
Kill switch & 30/30 & stops with zero extra settled ticks; explicit clear resumes to eval \\
Integrity & 80/80 & 70 corruptions fail closed; 10 torn tails recover \\
Output guard & 176/176 & protected-key, whitelist, catalog, numeric, cap, proposal cases \\
\bottomrule
\end{tabularx}
\end{table}

\paragraph{Interpretation.}
The utility ablation is intentionally a mechanism check, not a benchmark win: the simulator was calibrated so catalog-bounded round growth can reach the target. Its sharp separation verifies that the allowed path can matter without granting policy authority, while the idle null result prevents attributing benefit to every form of ``self-reflection.'' The fault results expose another important separation. Atomic commits make the projected scientific state reproducible under the tested deterministic planner, yet local atomic commits alone do not establish exactly-once external effects. Reporting 240/240 outcomes without the 210/240 trace result would hide that operational cost and side-effect risk.

\section{Responsible-use statement and limitations}

\paragraph{Responsible use.}
Meta-agent infrastructure is dual use: audit and recovery mechanisms can support responsible experimentation, but can also make autonomous optimization easier to operate at scale. Such systems could amplify a harmful goal, coordinate abusive tasks, centralize workplace decisions, or consume resources without meaningful consent. The main epistemic risk here is \emph{safety theater}---mistaking simulator invariants for deployment safety. We mitigate these risks by limiting scope and authority: there is no robot or over-the-air release path, no automatic release, and no mechanism that instantiates proposed agents. Machine-selected changes come from a fixed, code-owned catalog; goals, scope, evaluation identity, budget, and release thresholds remain human-controlled. Runs expose a kill path, proposal status, checkpoint provenance, hashes, and replayable accounting.

These controls do not remove operator responsibility or make a harmful human policy acceptable. Before real use, operators should assess affected people, independently cap monetary and energy spend, protect potentially sensitive log and task data with access and retention controls, authenticate approvals, externally anchor audit records, red-team prompt and tool interfaces, use a sealed holdout, and require domain-specific canary, rollback, and physical-safety review. Publishing boundary failures---including repeated calls and trace mismatches---is intended to support fault research, not certify autonomy.

\paragraph{Limitations.}
Evidence comes from one calibrated shirt-folding simulator, one simulated VLA, synthetic data, an offline deterministic planner, and simulated \gphs. Fifty seeds sample simulator noise, not model-provider, hardware, or organizational variation. The Feature catalog and dynamics were manually constructed; utility results validate the testbed mechanism rather than compare general learning algorithms. The same noisy evaluation drives feedback, stopping, and checkpoint selection, with no sealed holdout; latent score diagnoses winner's-curse bias but is unavailable in deployment. Exit uses a point estimate and regression detection is heuristic. The unsigned hash chain cannot resist a full-history rewrite, and approval identities are unauthenticated labels. Crash sites cover selected boundaries, not arbitrary storage, kernel, or network failures. Fixed settings are required for outcome equivalence; resource drift establishes only the accounting bound. A durable tick follows simulated work, so power loss could repeat real cost before recording. Finally, the protected-key list is duplicated across two enforcement modules, a maintenance hazard. The tests establish conformance to the modeled specification, not formal safety, security, or general self-improvement.

\section{Conclusion}

\sys{} makes bounded adaptation auditable through an explicit policy/evolution split, prefix compute accounting, and a recovery contract with two observables: scientific outcome and normalized execution trace. The experiments demonstrate a useful permitted growth path in the calibrated simulator and conformance at the tested boundaries. Their most consequential distinction is that all enumerated crashes recover the specified outcome while some repeat a planner call. A report of successful recovery should therefore state both what state was recovered and what execution was repeated. Extending this result to remote planners and physical workloads requires separate evaluation of nondeterministic retries, external side effects, authenticated state, and independently metered resource use.

\bigskip
\bibliographystyle{plainnat}
\bibliography{references}

\clearpage
\appendix

\section{Implementation details}

The implementation is a Python research testbed with a fixed orchestrator and five role-specific agents. Its bounded evolution state comprises seven optimization/data knobs---failure-replay boost, hard-task learning-rate multiplier, hard-task ceiling, minimum data-share floor, global learning-rate multiplier, synthetic-data ratio, and early-stop patience---plus a self-check flag. A model feature-pick output contains only a catalog ID, evidence, and rationale; code constructs the Feature and copies numeric deltas from the code-owned catalog. At most two Features may be merged per round and one during an idle pass. An agent-catalog output likewise contains an ID and rationale; code constructs the proposal with status \texttt{PENDING}. Neither a pending nor an approved record has an execution path in this prototype.

The event log is the source of truth; dashboards and candidate manifests are derived views. Before replay or append, the implementation verifies sequence numbers, predecessor hashes, and event hashes. Policy and evaluation files are separately verified against recorded hashes. Training snapshots and final checkpoints are content-address checked before use. Round growth is installed only by the durable \texttt{stage\_done} event, and idle growth only by \texttt{idle\_finished}; legacy fine-grained event types remain readable but are not emitted by new runs.

\section{Experimental protocol and extended results}

Experiments ran on Linux/x86-64 with Python 3.12.14. Efficacy and budget studies use seeds 0--49. Reliability studies use seeds 0--9. All planner responses use the offline deterministic rules; no network model is queried. The evaluation set has five equally weighted tasks with 200 episodes each. One evaluation costs 0.2 simulated \gphs. A default complete training round requests $4\times4=16$ \gphs{} plus evaluation, while the minimum affordable round is one $0.25$-hour step on four simulated GPUs plus evaluation, or 1.2 \gphs.

\begin{table}[h]
\caption{Point estimates for the full $2\times2$ Feature-growth ablation. Confidence intervals for the principal contrasts and attainment rates are reported in Section~\ref{sec:evaluation}.}
\label{tab:ablation-full}
\centering
\small
\begin{tabular}{@{}lrrrr@{}}
\toprule
Condition $(E,I)$ & Hits & RM \gphs{} & Selected obs. & Latent \\
\midrule
Full $(1,1)$       & 50/50 & 103.03 & 0.9606 & 0.9584 \\
Round-only $(1,0)$ & 50/50 & 103.03 & 0.9606 & 0.9584 \\
Idle-only $(0,1)$  & 0/50  & 160.00 & 0.8437 & 0.8324 \\
No-growth $(0,0)$  & 0/50  & 160.00 & 0.8437 & 0.8324 \\
\bottomrule
\end{tabular}
\end{table}

\begin{table}[h]
\caption{Budget sweep under the full loop. ``Spent'' is mean terminal accounted compute.}
\label{tab:budget-full}
\centering
\small
\begin{tabular}{@{}rrrrrr@{}}
\toprule
Budget & Hits & Spent & Selected obs. & Latent & Best $\ne$ last \\
\midrule
20  & 0/50  & 19.4  & 0.6223 & 0.6156 & 12/50 \\
40  & 0/50  & 39.6  & 0.7081 & 0.7082 & 1/50 \\
80  & 0/50  & 80.0  & 0.9202 & 0.9203 & 0/50 \\
160 & 50/50 & 103.0 & 0.9606 & 0.9584 & 0/50 \\
\bottomrule
\end{tabular}
\end{table}

\section{Fault model and interpretation}
\label{app:faults}

The crash comparison projects final state onto decision, accounted GPU and simulated time, ordered evaluation curve, checkpoint hashes, Feature identities and origins, proposal identities and status, and problem-list identities. Scientific traces omit wall-clock and latency fields plus expected recovery bookkeeping. Thirty trace mismatches occur only at \texttt{DECOMPOSE} pre-commit, \texttt{EVOLVE} pre-commit, and idle pre-commit (10 seeds each); each contains one additional offline-planner call. The final projection remains identical because that planner is deterministic and each state plan is committed atomically. A nondeterministic remote planner could return a different plan on retry.

The resource-drift suite starts at four GPUs with 0.25-hour steps, crashes after tick 1, and resumes through four paths: eight GPUs/0.25 hours; eight GPUs/0.5 hours; two GPUs/0.125 hours; and an eight-GPU/0.5-hour second crash followed by two GPUs/0.125 hours. All 40 runs keep both terminal and prefix usage within the 20-hour policy, retain monotone cumulative tick gauges, reconcile settlements, and complete evaluation. Trajectories are deliberately not compared with the fixed-config baseline.

Integrity cases mutate the policy file, evaluation file, checkpoint, snapshot, a complete ledger payload, a ledger hash, or delete a middle ledger entry; all 70 attack trials stop at a specified verification boundary. Ten partial final JSON records are treated as torn appends, truncated to the last complete prefix, and successfully resume. This matrix models localized corruption. With complete write access, an attacker can replace policy, evaluation, artifacts, and the full ledger while recomputing hashes; signatures, authentication, trusted timestamps, and external anchoring are future requirements.

\section{Reproducibility artifacts}

The supplementary materials described in this study comprise the experiment driver, figure generator, compact per-trial CSV files, and a machine-readable summary. These experimental scripts, per-trial files, and raw run directories are not included in this preprint's LaTeX source package. The driver rejects a raw-data root containing an existing study manifest or measurements by default, records source hashes and its protocol in a manifest, uses independent run directories per seed and condition, and fails on incomplete paired blocks. The codebase passes 532 unit and integration tests after the reliability fixes evaluated here.

\section{Conditional reasoning about run contracts}
\label{app:analysis}

The following arguments make the assumptions behind the accounting and recovery checks explicit. They are consequences of the stated execution model, not additional experiments or claims that the Python implementation has been formally verified.

\paragraph{Safe admission after repricing.}
At a repricing boundary, let $H$ be the already accrued recorded charge, including completed in-flight ticks, and let $B$ be the policy cap. Let $e\geq0$ be the remaining evaluation charge not yet included in $H$, and let every additional training tick under the current resource configuration have known charge $r>0$. Assume all relevant charges are recorded, no other new charges occur during this allocation, no concurrent unreserved work is admitted, and evaluation costs at most $e$. When $H+e\leq B$, an admissible number of further ticks is bounded by
\begin{equation}
 n\leq n_{\max}=\left\lfloor\frac{B-H-e}{r}\right\rfloor.
 \label{eq:admission}
\end{equation}
When $H+e>B$, the combined training-and-evaluation sequence cannot be admitted under these assumptions; clipping a negative numerator to zero does not make the evaluation affordable.

For each intermediate tick count $0\leq j\leq n$, we have
$H+jr+e\leq H+nr+e\leq B$ by Equation~\ref{eq:admission}.
Since $e\geq0$, the prefix charge before evaluation also remains within the cap. A pure bookkeeping transfer of charge $q$ maps $(H_s,H_i)$ to $(H_s+q,H_i-q)$ and preserves their sum. A settlement event that adds a new charge is not such a pure transfer and must account for that charge separately. Reapplying the same admission check to the current $H$, $e$, and $r$ at each resource change yields the corresponding bound across repricing boundaries by induction, under the same recording and serialization assumptions. The implementation uses a numerical tolerance; this exact-arithmetic argument does not analyze floating-point accumulation or unrecorded physical cost.

This argument explains the role of evaluation reservation and cumulative charges: changing GPUs or tick duration changes the cost of future work, not the policy cap or the charge already recorded. It also identifies assumptions that must be reconsidered for concurrent workers, uncertain physical costs, or a delayed charge meter.

\paragraph{Why atomic state commits do not ensure repeat-free effects.}
Consider a non-idempotent external operation whose completion is not atomic with its local durable commit. After a crash removes volatile knowledge, two histories can expose the same local ledger prefix: in one, the external effect occurred but its receipt was not committed; in the other, the operation did not take effect. Assume no external protocol reliably resolves or deduplicates the operation, such as an idempotency mechanism, transactional coordination, or conclusive durable-outcome query. A recovery rule based only on the common local prefix cannot distinguish the histories. Retrying can duplicate the effect in the first history; declining to retry can leave the operation incomplete in the second. Hence local commit atomicity alone cannot guarantee both eventual completion and at-most-once external effects under these assumptions.

This is a standard limitation of the execution boundary, not a claim of a new impossibility result. It explains why the record/replay perspective~\citep{burckhardt2021durable} requires a precisely stated effect model. In our offline experiments, the repeated operation is a deterministic planner call, so it can leave the scientific-state projection unchanged. An external service with side effects would require additional mechanisms and evidence that are not provided by these experiments.

\section*{Use of generative AI tools}
OpenAI Codex assisted with manuscript restructuring, language revision, formalization of stated assumptions, reference checks, and LaTeX preparation. The numerical experiment results are from the underlying study; no new experimental runs were performed for this revision.

\end{document}